\documentclass[preprint,12pt,authoryear]{article}
\usepackage{graphicx} 
\usepackage{booktabs} 
\usepackage{multirow} 
\usepackage{booktabs} 
\usepackage{amsmath}

\title{Beyond Black Boxes: Visual Grounding for Text Embeddings}
\author{Slimane Larabi\\
Computer Science Faculty, USTHB University}
\date{August 2026}

\begin{document}
\maketitle
\begin{abstract}
In this paper, our goal is to represent text embeddings as sketches for three main reasons. First, to visualize text representation and to explore their potential for AI explainability. Second, to investigate the potential benefits of the sketch-based representation for information retrieval. Third, to explore their potential for reasoning, drawing inspiration from the concept of mental imagery.
We propose a sketch-based visual grounding framework that transforms text representations into visual topographic sketches. Our approach aims to emphasis the semantics content encoded in the text embeddings through the corresponding sketch. 

Evaluations are conducted on the MS-COCO dataset to examine the utility of these sketches for text retrieval. The comparative study of SBERT and Sketch-based method reveals that SBERT retains an advantage in strict top-1 retrieval accuracy while achieving comparable performance  to the sketch-based method in top-5 retrieval. A cascaded approach combining SBERT and the sketch-based method outperforms SBERT alone, suggesting that the Sketch-based representation could offer significant potential for information retrieval if trained directly on text.
\end{abstract}

\section{Introduction}
\label{sec:intro}

Recent computer vision models and text embedding frameworks, such as Sentence-BERT (SBERT) ~\cite{Reimers2019}, employ high-dimensional vector spaces to represent the semantics of images, videos, and their captions. To generate this fixed-size representation, SBERT, for instance, adapts pre-trained Transformer architectures using Siamese and triplet network structures. These representations enables efficient assessment of semantic similarity using cosine similarity. However, while these representations excel at their respective tasks, they remain difficult to interpret that hinders efforts to improve model explainability.

Fixed-size embeddings can lose fine-grained semantic information as text length increases. They may be also sensitive to minor natural variations or adversarial perturbations ~\cite{luan-etal-2021} ~\cite{Dinzinger2025} ~\cite{Asl2023}. It is also worth noting their difficulty in conveying granular semantic details or capturing long-range dependencies ~\cite{chen-etal-2024}~\cite{Jin2025}.
To the best of our knowledge, standard text encoders have not been used to project linguistic descriptions onto topographic sketches. In the present work, the sketch is derived from the SBERT vector; consequently, these limitations are inherited rather than resolved. Investigating these aspects will be the subject of future work, in which the sketch would be generated directly from the text by an end-to-end trained model, rather than from SBERT representations.
The fundamental challenge, therefore, lies in bridging the gap between abstract mathematical representations and intuitive human understanding. This requires a mechanism capable of transforming opaque vector spaces into explicit, visually analyzable artifacts (such as sketches) that preserve rich semantic relationships while offering greater transparency.

\begin{figure}
 \centering
    \includegraphics[width=0.95\linewidth]{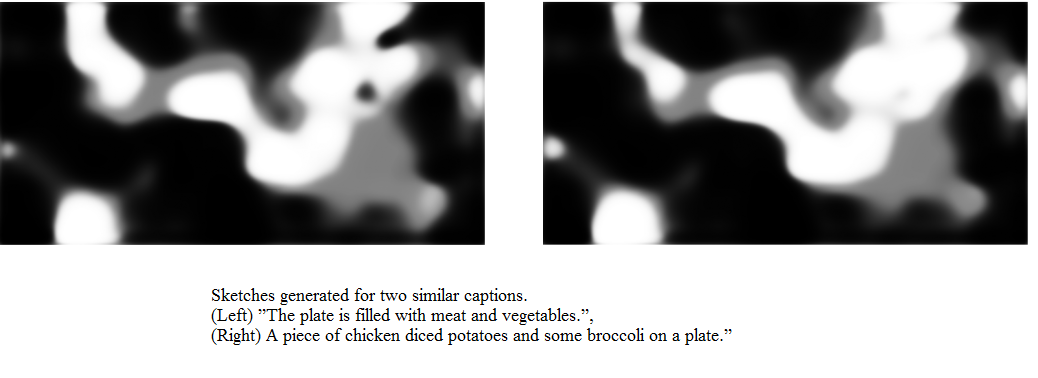}
    \caption{Sketches generated for two similar captions.(Left) "The plate is filled with meat and vegetables.", (Right) A piece of chicken diced potatoes and some broccoli on a plate."}
    \label{fig:1}
\end{figure}

In this paper, we propose a novel framework that transforms the vector representations of a sentence (e.g. an image or video caption) into structured topographic sketch. Figure \ref{fig:1} illustrates sketches generated for two captions describing the same visual content. Unlike conventional text-embedding models, our approach provides an explicit visual representation of the embedding, which can be used to improve the SBERT performance in various applications (e.g. sentence retrieval, reasoning). Our comprehensive evaluations reveal important performance trade-offs: while pure SBERT maintains an advantage in strict top-1 precision, the sketch-based representation closely approaches SBERT performance at top-5 retrieval under comparable settings and combining SBERT and Sketch improves overall performance.   

This paper is organized as follows: Section 2 reviews SBERT embeddings. Section 3 presents our framework for mapping SBERT caption embeddings to topographic sketches via joint contrastive and spatial regularization. Section 4 provides an extensive evaluation and comparative study of sketch representations against standard SBERT. Finally, Section 5 concludes with a summary of finding and future perspectives.   

\section{Review of SBERT}

Sentence-BERT (SBERT) ~\cite{Reimers2019} is a modification of the pretrained BERT network that uses Siamese and triplet network structures to derive semantically meaningful sentence embeddings. Before its introduction, using standard BERT for tasks like semantic textual similarity (STS) or clustering required feeding sentence pairs simultaneously into a cross-encoder. This results in prohibitive computational costs for large datasets: finding the most similar pair among $n$ sentences requires ($O(n^2)$ transformer forward passes. SBERT removes this bottleneck by turning BERT into a bi-encoder.  
SBERT passes sentences independently through two identical BERT networks with shared weights. To map variable-length token outputs into a fixed-size dense vector, it applies pooling operations, 
 with mean pooling found empirically to work best. Because sentences are encoded independently, embeddings can be precomputed and indexed. The number of transformer passes drops from $O(n^2)$ to $O(n)$ and similarity search relay on fast vector operations such as cosine similarity.  

During fine-tuning, SBERT relies on specialized objective functions to structure the vector space for semantic proximity:\\

\textbf{Classification Objective}: When fine-tuning on Natural Language Inference (NLI) datasets, where sentence pairs are labeled as entailment, contradiction, or neutral, SBERT concatenatnes the two sentence embeddings with their element-wise absolute difference and passes this combined representation to a softmax classifier. This encourages the underlying transformer to learn features that distinguish semantic relationships and logical entailment.\\

\textbf{Regression Objective}: When the training data provides graded semantic similarity scores (such as human-annotated ratings normalized to $[0,1]$, SBERT is trained with a mean squared error loss (MSE). The model calculates the cosine similarity between two sentence embeddings and adjusts its weights to minimize the difference between this predicted similarity score and the human-annotated ground truth.\\

\textbf{Triplet objective}: This objective use an "anchor" sentence, a "positive" sentence, and a "negative" sentence. The loss function tunes the vector space using a margin-based approach: it pulls the anchor and positive embeddings closer together while actively pushing the anchor and negative embeddings farther apart, ensuring a distinct geometric separation between different meanings.  

\section{Mapping SBERT Embeddings to Sketches}

Our objective is to map the vector embedding of a caption, encoded by Sentence-BERT (SBERT), into a corresponding visual sketch. This process is conceptually inspired by human cognition: humans naturally form a mental image of a given idea or sentence to reason about it. 

Motivated by this ability to construct mental representations with regions corresponding to specific objects and actions, our goals are threefold: (1) to visualize the embedding vector, allowing us to inspect what semantic information is captured and how the sketch varies under caption perturbations; (2) to assess whether the resulting sketch can serve as an effective representation for information retrieval; and (3) to investigate how the sketch can contribute to reasoning when used as an imaginary image~\cite{Larabi_IJCLR2026}. Only the first two objectives are investigated in this paper, while the third will be processed as future work.

To achieve this cross-modal mapping, our architecture relies on an end-to-end pipeline combining a semantic text encoder, a topographical decoder, and a visual feature re-encoder, optimized jointly via contrastive and spatial regularization losses.

\subsection{Semantic Text Encoding}

The input caption is first transformed into a dense 384-dimensional vector using a pretrained SBERT model (\textit{all-MiniLM-L6-v2}) \cite{Reimers2019}. To introduce non-linearity, the 384-dimensional vector is processed by a multilayer perceptron (MLP) composed of a linear layer with a 256-dimensional hidden layer, a ReLU activation, and a dropout layer (rate $0.1$) that reduces overfitting to specific training captions. A second linear layer produces the 384-dimensional latent vector $z$, which is standardized by layer normalization to stabilize its scale before it is passed to the sketch decoder.

\subsection{Topographical Sketch Generation (Decoder)}

The latent vector $z$ is fed directly into the sketch decoder. This module maps abstract text semantics onto a spatial canvas: a fully connected layer first expands $z$ into a base tensor of size $128 \times 13 \times 25$.

To progressively increase the spatial resolution, the decoder uses a sequence of three upsampling blocks. Each block combines batch normalization, a ReLU activation, bilinear upsampling with a scale factor of $2$, and a $3 \times 3$ convolution (reducing the channels from $128$ to $64$, $32$, and finally $1$). Upsampling followed by convolution, instead of transposed convolutions, is commonly used to reduce checkerboard artifacts. A final bilinear interpolation then resizes the tensor to a fixed resolution of $416 \times 800$ pixels with a single channel ($C=1$).

A concluding sigmoid activation bounds the pixel intensities between $0$ and $1$, producing a grayscale density map.

\subsection{Visual Feature Re-Encoding}

The generated sketch is then processed by a convolutional sketch encoder, applied after the decoder. It extracts visual features through three convolutional layers ($32$, $64$, and $128$ channels, kernel size $4$, stride $2$, padding $1$), each followed by a ReLU activation. An adaptive average pooling layer reduces the feature map to a fixed $4 \times 8$ grid, and flattening followed by a linear projection maps the result to a 384-dimensional feature vector, which is used for retrieval.

\subsection{Joint Optimization Objective}

The network is optimized end-to-end with a hybrid objective combining a contrastive alignment loss and a spatial variation loss. At each step, we sample $B=32$ distinct groups of captions and, for each group $i$, two different captions: an anchor $a_i$ and a positive $p_i$. Each caption embedding $x$ is passed through the pipeline to obtain a latent vector $z$, a sketch $S \in [0,1]^{416 \times 800}$, and a feature vector $f$ from the sketch encoder, which is $\ell_2$-normalized as $\tilde f = f/\lVert f \rVert_2$.

\paragraph{Contrastive alignment.}
Alignment is enforced with an InfoNCE loss computed on $\ell_2$-normalized sketch features. InfoNCE (Information Noise-Contrastive Estimation) was introduced in ~\cite{oord2019} as a contrastive loss inspired by Noise-Contrastive Estimation (NCE). It maximizes a lower bound on the mutual information between related views of the data by training the model to identify the positive pair among a set of negative distractors. In our setting, captions describing the same underlying scene (for example, two different phrasings of the same content) are positives and should map to similar features, while unrelated captions are negatives. The similarities are scaled by a temperature parameter, set to $0.07$, so that matching captions are pulled together and unrelated ones are pushed apart.

\begin{equation}
\mathcal{L}_{\mathrm{NCE}} = \frac{1}{2B}\sum_{i=1}^{B}\left[
-\log\frac{\exp(\tilde f_{a_i}^{\top}\tilde f_{p_i}/\tau)}{\sum_{j=1}^{B}\exp(\tilde f_{a_i}^{\top}\tilde f_{p_j}/\tau)}
-\log\frac{\exp(\tilde f_{a_i}^{\top}\tilde f_{p_i}/\tau)}{\sum_{j=1}^{B}\exp(\tilde f_{a_j}^{\top}\tilde f_{p_i}/\tau)}
\right].
\end{equation}

Where $B=32$ is the number of caption groups in a batch; $a_i$ and $p_i$ are two different captions randomly drawn from group $i$ (the anchor and the positive), both describing the same scene; $\tilde f_{a_i}$ and $\tilde f_{p_i}$ are their $\ell_2$-normalized 384-dimensional features produced by the sketch encoder, so that $\tilde f_{a_i}^{\top}\tilde f_{p_j}$ is a cosine similarity; and $\tau=0.07$ is a fixed temperature that controls the sharpness of the softmax. For each anchor $a_i$, the only positive is $p_i$, whereas the $B-1$ captions $p_j$ ($j\neq i$) drawn from the other groups of the batch serve as negatives; the second term of the loss plays the same role with the roles of anchors and positives exchanged. Minimizing $\mathcal{L}_{\mathrm{NCE}}$ therefore brings the sketch features of captions of the same group closer together and pushes apart those of different groups, so that the sketches encode the semantic content of the caption.

\paragraph{Spatial variation.}
To avoid the trivial solution of a uniform or blank canvas, which is easy to reach early in training and from which the network may not recover, we reward spatial variation in the generated sketches. Let $G$ denote a $5\times5$ average filter (stride 1, zero padding excluded from the average). The loss is the negative spatial variance of the smoothed sketch, averaged over the batch:
\begin{equation}
\mathcal{L}_{\mathrm{var}}(S) = -\frac{1}{B}\sum_{b=1}^{B}\operatorname{Var}_{h,w}\!\big[(G * S_b)(h,w)\big].
\end{equation}
Where:
\begin{itemize}
    \item \textbf{$S_b$:} The $b$-th sketch of the batch, represented as a grayscale image of dimensions $416 \times 800$ pixels with intensity values constrained between $0$ (black) and $1$ (white) via a Sigmoid activation function, with a total batch size of $B = 32$ sketches. \textbf{$G * S_b$:} The smoothed sketch obtained via a $5 \times 5$ averaging filter $G$ (where all coefficients equal $1/25$). Each pixel is replaced by the local neighborhood average ensures that border handling excludes padding zeros).
    \item \textbf{$\operatorname{Var}_{h,w}[\cdot]$:} The spatial variance computed across the pixel values of a single image, where $h$ and $w$ index the rows and columns:
    $$\operatorname{Var}_{h,w}[x] = \frac{1}{HW}\sum_{h,w}\big(x(h,w)-\bar x\big)^2$$
    with $\bar x$ representing the mean pixel value. This quantifies image contrast: it is zero for uniform images and large when distinct dark and bright regions coexist.
\end{itemize}

\paragraph{Total loss.}
\begin{equation}
\mathcal{L} = \mathcal{L}_{\mathrm{NCE}} + \lambda\left[\mathcal{L}_{\mathrm{var}}(S_{a}) + \mathcal{L}_{\mathrm{var}}(S_{p})\right],\qquad \lambda = 0.01,
\end{equation}
where $S_a$ and $S_p$ are the sketches generated from the anchor and positive captions. We train with AdamW (learning rate $10^{-3}$) and a cosine annealing schedule for a fixed number of steps.

\section{Experimental Evaluation}
\label{sec:experiments}

\subsection{Semantic Information Coded by Sketches}

\begin{figure}
 \centering
    \includegraphics[width=0.3\linewidth]{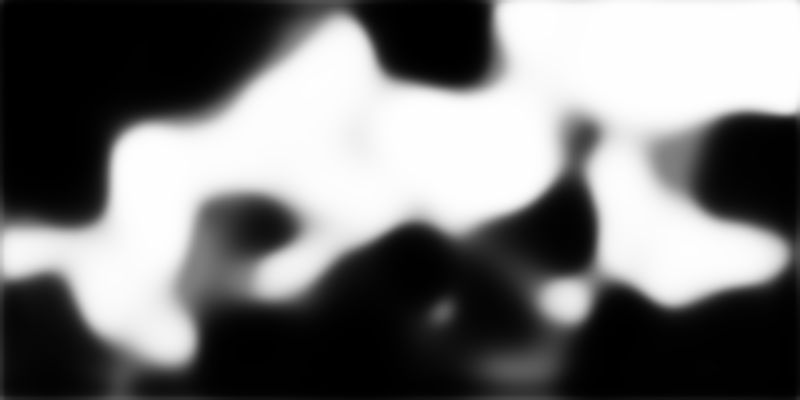}
    \includegraphics[width=0.3\linewidth]{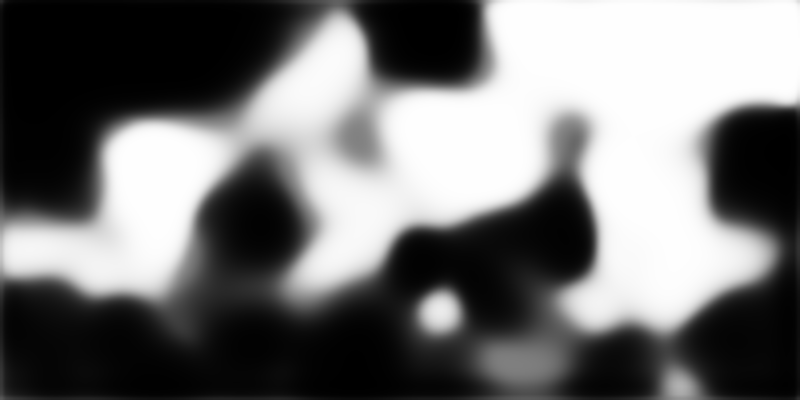}
    \includegraphics[width=0.3\linewidth]{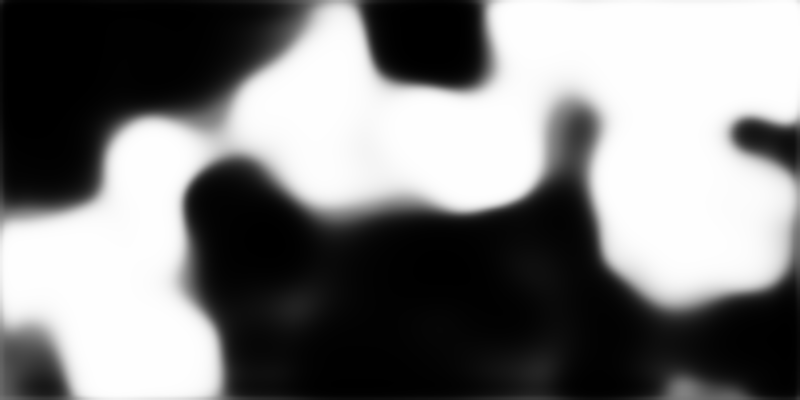}
    \caption{Sketches generated for the captions "a man is running in house", "a man is running in the garden", "a man is running in the street"}
    \label{fig:2}
\end{figure}

\begin{figure}
 \centering
    \includegraphics[width=0.3\linewidth]{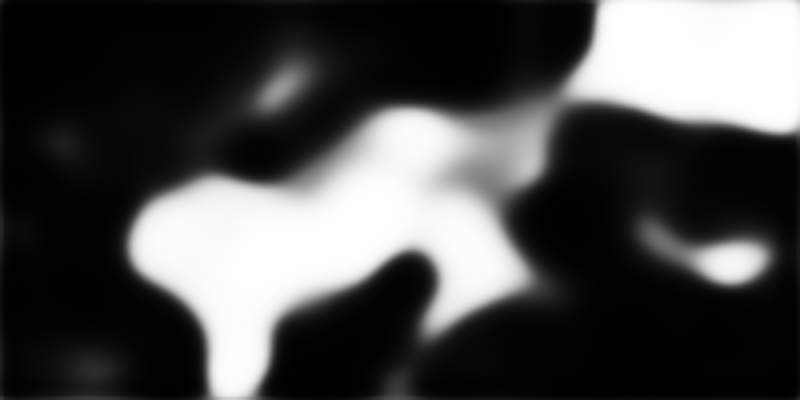}
    \includegraphics[width=0.3\linewidth]{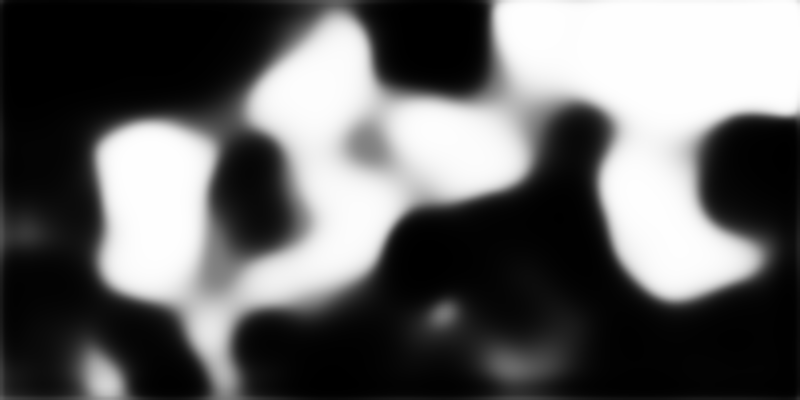}
    \includegraphics[width=0.3\linewidth]{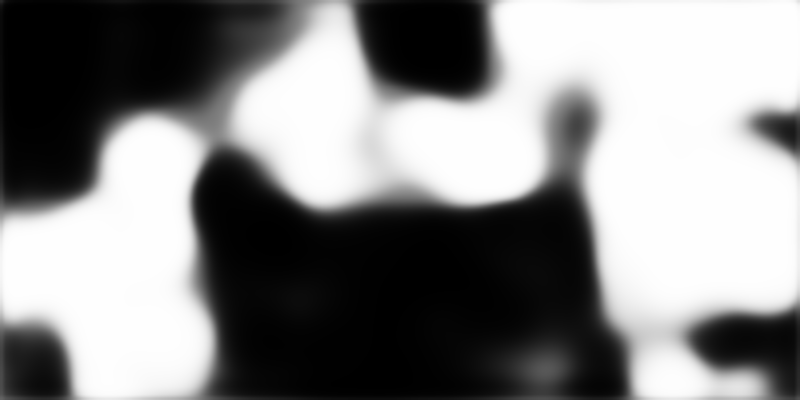}
    \caption{Sketches generated for the captions "a girl is running in house", 
    "a boy is running in the garden", "An elderly man is running down the street."}
    \label{fig:2p}
\end{figure}

Our model was trained on the MS-COCO dataset ~\cite{MSCOCO2014}. 
Figure \ref{fig:2} shows the sketches generated for the captions "man is running in house", "man is running in the garden", "man is running in the street". The sketches share common information corresponding to "man is running" together with variable information related to the location. When the word "man" is replaced with "girl", "boy" or elderly man", we obtain sketches shown in figure \ref{fig:2p}. Some differences appear, but the sketch keep the global layout, which suggests that the model encodes common notion of a person.

\begin{figure}
 \centering
    \includegraphics[width=0.3\linewidth]{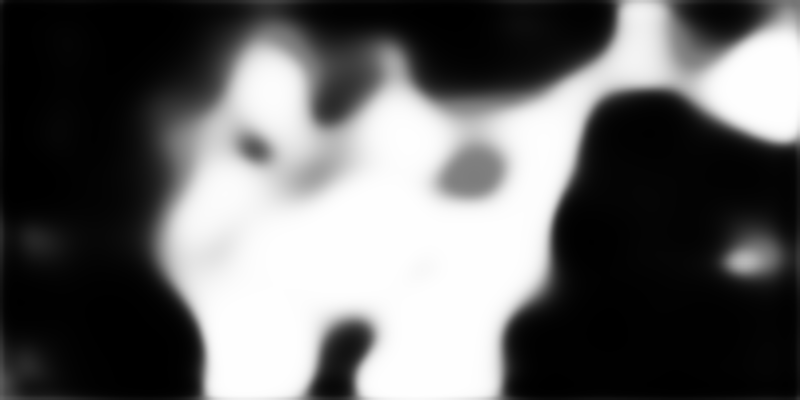}
    \includegraphics[width=0.3\linewidth]{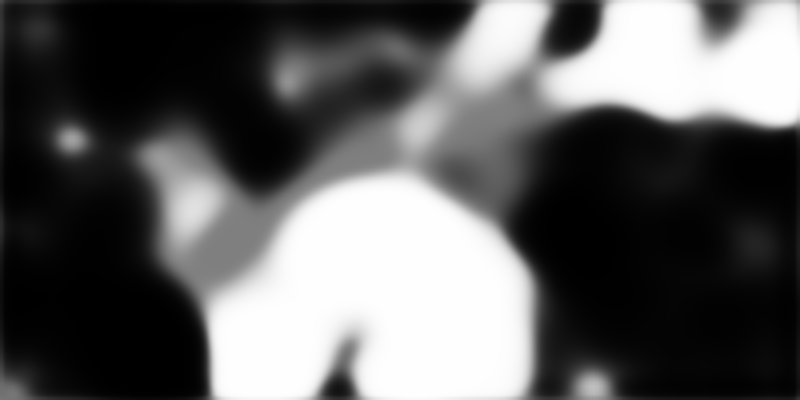}
    \includegraphics[width=0.3\linewidth]{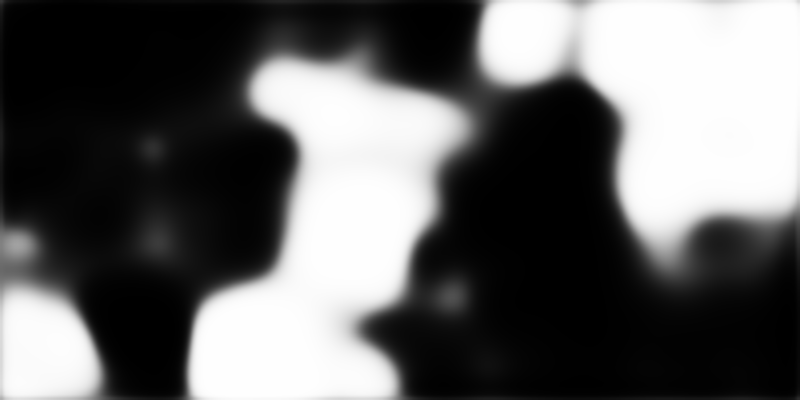}
    \caption{Sketches generated for the captions "a dog is running in house", 
    "a dog is running in the garden", "a dog is running down the street."}
    \label{fig:3}
\end{figure}

Changing the subject (dog instead of man), yields sketches that are visually different from the previous ones. Figure \ref{fig:3} shows the sketches generated from the captions "a dog is running in house", "a dog is running in the garden" and " a dog is running in the street".

\begin{figure}
    \centering
    \includegraphics[width=0.3\linewidth]{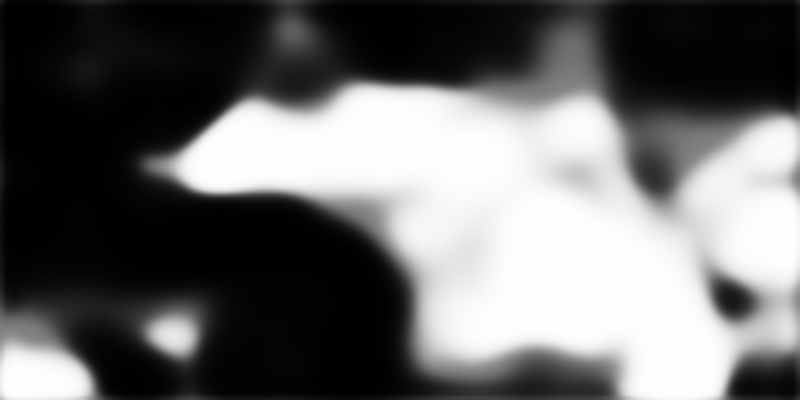}
    \includegraphics[width=0.3\linewidth]{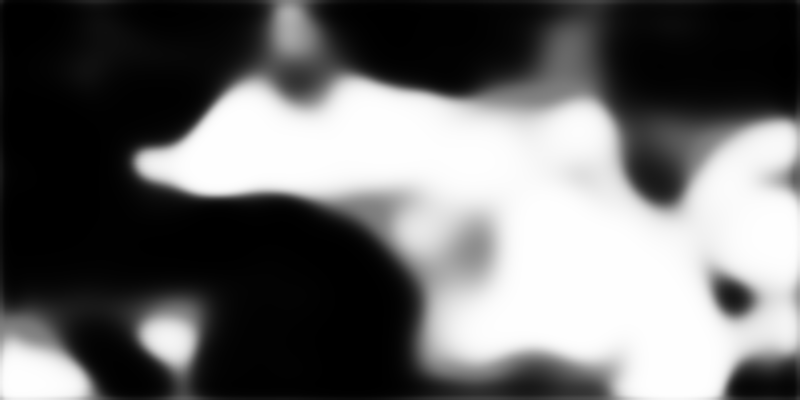}
    \includegraphics[width=0.3\linewidth]{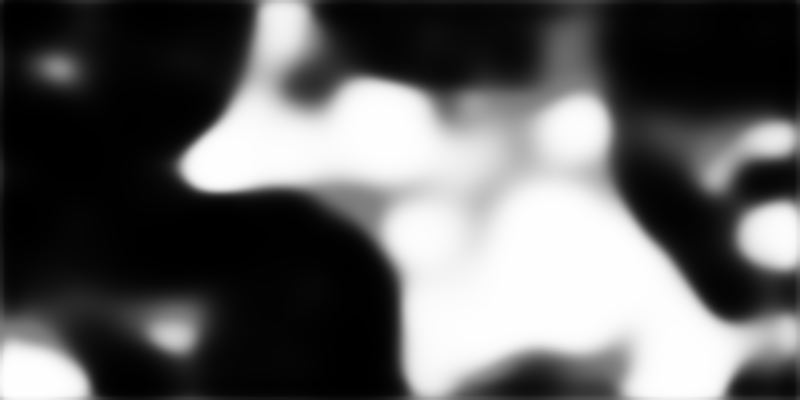}
    \includegraphics[width=0.3\linewidth]{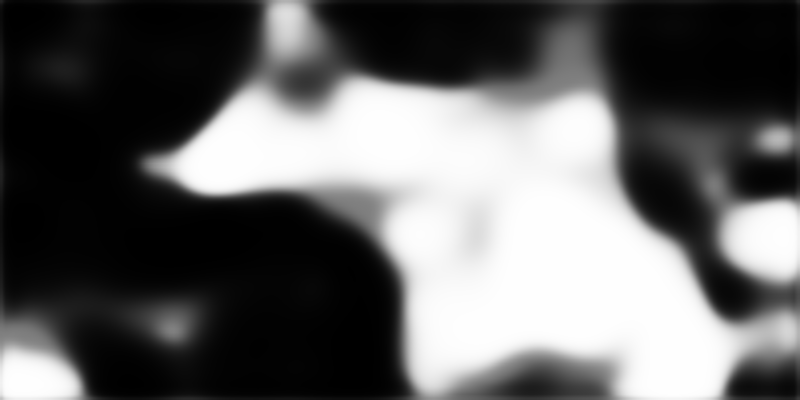}
    \includegraphics[width=0.3\linewidth]{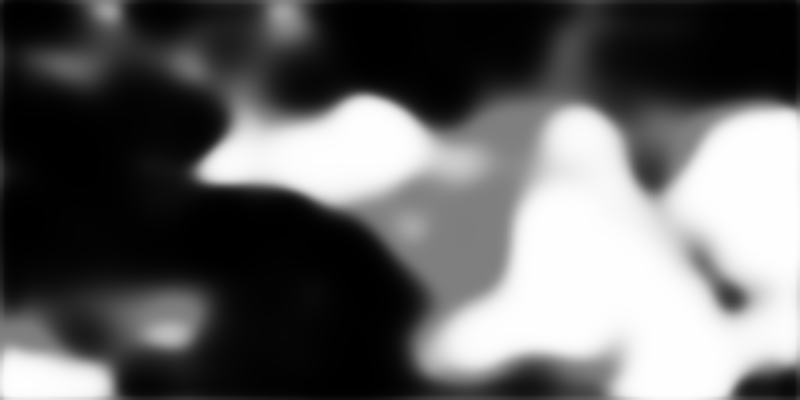}
    \caption{Sketches generated for the captions "A white plate topped with meat and vegetables.", "A close-up of a plate of food with meat and veggies.",  "A meal consisting of meat, broccoli, and other vegetables on a plate.", "A plate with meat and green vegetables on it.", "Some food that is on a white plate."}
    \label{fig:4}
\end{figure}

Finally, we took five captions of the same image from MS-COCO dataset and generated the corresponding sketches (Figure \ref{fig:4}). The sketches are visibly similar despite the difference in the words used.

\subsection{Sketch Variation Under Caption Perturbation}

Many tests have been performed to study the sketch variation under caption perturbation. As example, to the original caption: "A golden retriever catches a red frisbee in mid-air at the public park.", we got for each kind of perturbations the followings captions:
Lexical Perturbation: "A golden retriever catches a crimson flying disc mid-air at the public park.",\\
Syntactic Reordering: "At the public park, a golden retriever catches a red frisbee in mid-air.",\\
Typo and Noise Perturbation: "A golde retriver catches a red frisbee in mid-air at the prk.",\\
Partial omission: "A golden retriever catches a red frisbee in mid-air." \\

To measure similarity between sketches, the Sketch encoder extracts high-dimensional feature vectors (384 dimensions) from each rendered output. These extracted feature representations are then normalized using L2 normalization along their feature dimension. Finally, the similarity between a reference sketch and comparative sketches is quantified by computing the cosine similarity through matrix multiplication of their normalized feature vectors.

\begin{figure}
    \centering
    \includegraphics[width=0.5\linewidth]{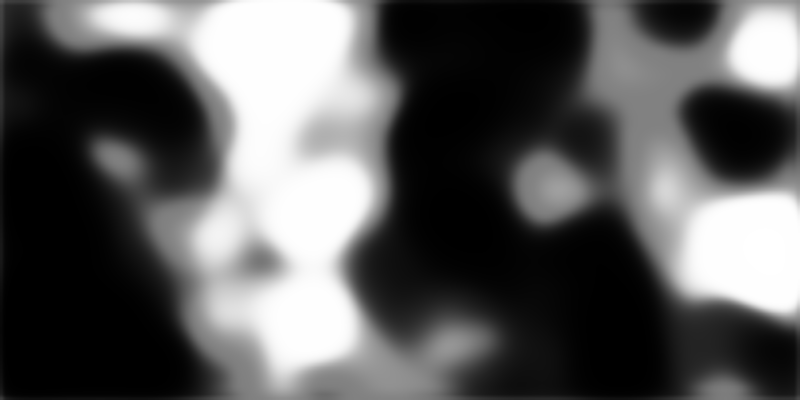}\\
    \includegraphics[width=0.48\linewidth]{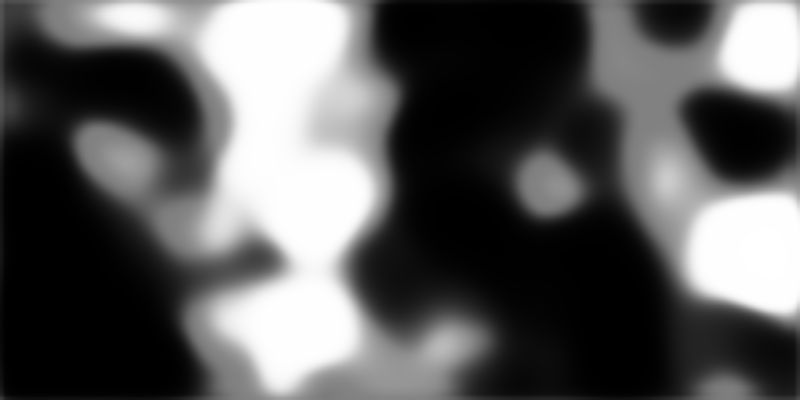}
    \includegraphics[width=0.48\linewidth]{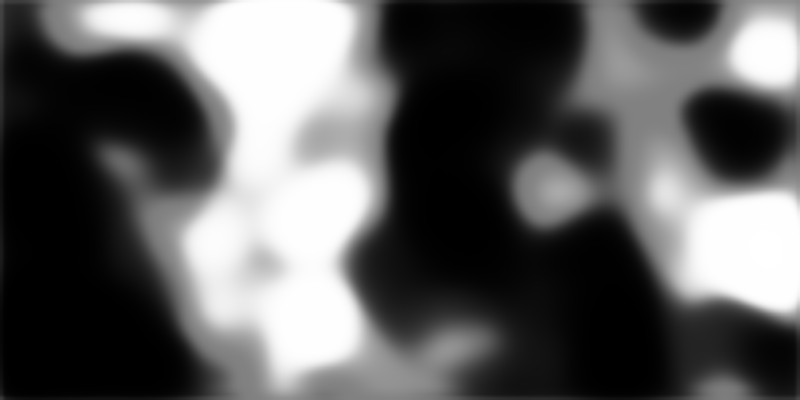}
    \includegraphics[width=0.48\linewidth]{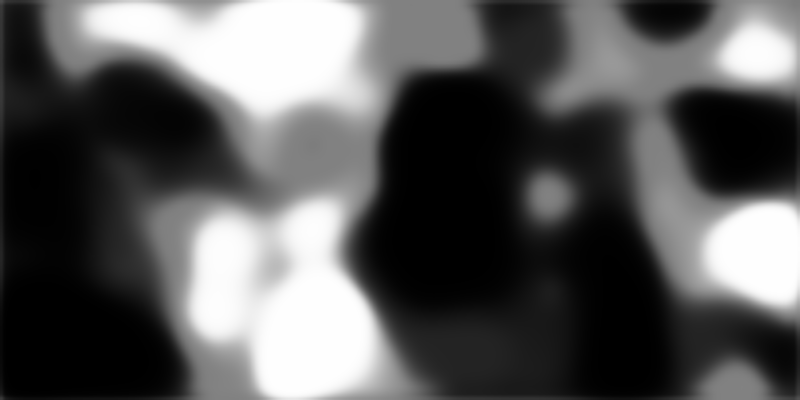}
    \includegraphics[width=0.48\linewidth]{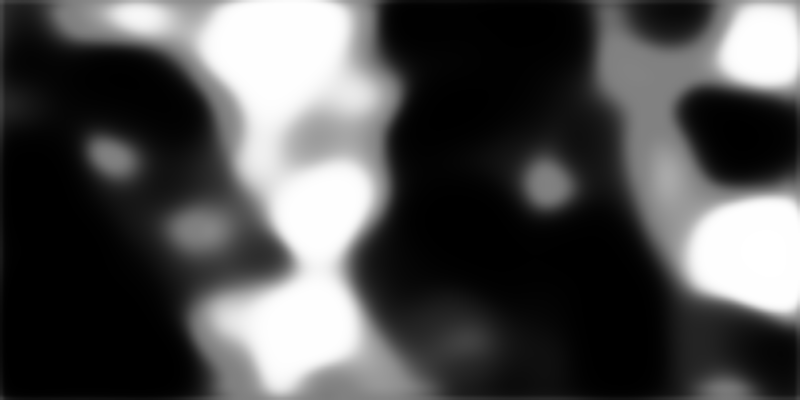}
    \caption{Sketches generated for the caption original and respectively for a Lexical perturbation, Syntactic Reordering, Typo and Noise Perturbation, Partial Omission.}
    \label{fig:5}
\end{figure}

The results of similarity measure between original and four perturbed captions (Figure \ref{fig:5}) demonstrate that the Sketch feature space remains robust and visually consistent. Syntactic reordering preserves highest fidelity, whereas character-level typos induce the most degradation. Overall, the visual decoder acts as an abstraction layer that filters out word-level variations to maintain spatial and structural consistency.
The similarities scores are equal respectively to : 0.9763, 0.9920, 0.8405, 0.9581. Note that for a caption which is different from the original: "A sleek black cat leaps gracefully across a rustic wooden fence at twilight." the computed similarity between their sketches is equal to: 0.2356.

\subsection{Sketch vs. SBERT Caption-to-Caption Retrieval}

We evaluated our sketch-based retrieval method against a pure SBERT baseline on caption retrieval over MS-COCO. We first describe the setup, then report the results and analyze the effect of the training set size and of the gallery size.

\subsubsection{Experimental Setup}
We use the MS-COCO captions~\cite{MSCOCO2014} where each item is a group contains up to five captions describing the same image. Only groups with at least three captions are kept. The sketch pipeline is trained exclusively on MS-COCO on subsets of $1{,}000$, $10{,}000$, $20{,}000$ and $30{,}000$ caption groups.

\paragraph{Retrieval protocol.}
For each evaluation, we randomly sample $N \in \{500, 1000, 5000\}$ groups (for fixed seed $42$). From each group, one caption is randomly held out as the \emph{query}, and the remaining captions of all $N$ groups form the \emph{gallery} (at most $4N$ captions).A query is ranked against every gallery caption by cosine similarity, and a retrieved caption is counted as correct if it belongs to the same group as the query. We report \textbf{Precision@1} (the top-ranked caption is correct) and \textbf{Precision@5} (at least one of the five top-ranked captions is correct). Because each evaluation uses $N$ queries against a gallery built from the same $N$ groups, increasing $N$ increases both the number of queries and the number of distractors.

\paragraph{Methods.}
\textbf{SBERT (pure)} embeds captions with the pre-trained \texttt{all-MiniLM-L6-v2} model ($384$-dimensional) and ranks by cosine similarity; it requires no training and is therefore identical across training-set sizes. \textbf{Ours (Sketches)} maps a caption's SBERT embedding through a semantic encoder (MLP) to a latent vector, decodes it into a $416\times 800$ grayscale sketch, and re-encodes the sketch with a convolutional encoder into a $384$-dimensional feature vector; retrieval is performed by cosine similarity in this feature space, for both queries and gallery captions.

\subsubsection{Main Results}
Table~\ref{tab:main_results} reports the results for all settings.
\begin{table*}[t]
\centering
\caption{Caption-to-caption retrieval precision (\%) on MS-COCO. $N$ is the number of evaluated groups, which equals the number of queries; the gallery contains the remaining captions of these $N$ groups. Sketches are trained on $1$k, $10$k, $20$k or $30$k groups. The best sketch result in each row is in bold.} Training on more groups improves the sketches consistently, for every $N$ and both metrics. The largest gain occurs between $1$k and $10$k groups: for $N=500$, P@1 rises from $33.6\%$ to $51.0\%$ ($+17.4$ points) and P@5 from $60.0\%$ to $81.2\%$ ($+21.2$ points). Going from $10$k to $30$k groups brings improvements.

\label{tab:main_results}
\begin{tabular}{llccccc}
\toprule
\multirow{2}{*}{$N$ (queries)} & \multirow{2}{*}{Metric} & \multirow{2}{*}{SBERT (pure)} & \multicolumn{3}{c}{Ours (Sketches), training groups} \\
\cmidrule(lr){4-7}
 & & & 1k & 10k & 20k & 30k \\
\midrule
\multirow{2}{*}{500}  & P@1 & 67.20 & 33.60 & 51.00 & 53.80 &  \textbf{61.60} \\
                      & P@5 & 86.20 & 60.00 & 81.20 & 85.20 & \textbf{84.20} \\
\midrule
\multirow{2}{*}{1000} & P@1 & 55.60 & 21.60 & 37.60 & 44.30 & \textbf{45.70} \\
                      & P@5 & 80.80 & 43.50 & 69.70 & 74.50 & \textbf{75.70} \\
\midrule
\multirow{2}{*}{5000} & P@1 & 36.42 & 9.74  & 18.80 & 22.80 & \textbf{24.04}\\
                      & P@5 & 58.70 & 23.40 & 41.42 & 46.34 & \textbf{47.90} \\
\bottomrule
\end{tabular}
\end{table*}

SBERT remains the strongest method in every configuration. The gap depends strongly on the metric.
The sketch features therefore place the correct group among the top candidates almost as reliably as SBERT, while ranking the exact best match less precisely.


\subsection{Evaluation of SBERT and Skertch in cascade}

\paragraph{Data and protocol.}
We use the COCO Captions quintets dataset, in which each group contains five captions describing the same image. For each run, we sample 5{,}000 groups. In every group, two captions are held out as queries (10{,}000 queries in total) and the remaining three form the retrieval pool (about 15{,}000 captions). A retrieved caption is counted as correct only if it belongs to the same group as the query; this ground truth is strict, since captions of other images may be semantically valid matches, so absolute scores are underestimated. The groups are split into a validation part
(30\%, 3{,}000 queries) used to select all hyperparameters, and a disjoint test part (70\%, 7{,}000 queries) used for reporting. The whole procedure is repeated with five random seeds (42--46), each seed drawing a different set of groups
and a different split.

\paragraph{Representations.}
Captions are embedded with SBERT (\texttt{all-MiniLM-L6-v2}, 384 dimensions). The sketch model, trained beforehand on 30{,}000 COCO groups, maps an SBERT embedding to a latent vector $z$ (semantic encoder), decodes it into a $416{\times}800$ topographical sketch, and re-encodes the sketch into a 384-dimensional feature vector. We compare:\\
(i)~\emph{SBERT}; (ii)~\emph{Sketch}, retrieval with the sketch features; (iii)~\emph{Latent}, retrieval with $z$, i.e.\
the semantic encoder alone, without going through the image (ablation); (iv)~\emph{Untrained sketch}, the same
architecture with random weights (control); and cosine similarity is used throughout.

\paragraph{Combination methods.}
\emph{Score fusion} ranks pool captions by $s = \alpha\, s_{\mathrm{SBERT}} + (1-\alpha)\, s_{\mathrm{other}}$, where
$s_{\mathrm{other}}$ is the Sketch, Latent or untrained-sketch cosine score. Scores are used either raw or z-normalized per query; $\alpha \in \{0, 0.05, \dots, 1\}$. The mode and $\alpha$ are chosen on the validation split by maximizing the mean of Top-1 and Top-5 (ties favor the larger $\alpha$, i.e.\ SBERT). A \emph{threshold cascade} keeps the SBERT ranking when the best SBERT similarity exceeds a threshold and falls back to the Sketch ranking otherwise (threshold also selected on validation). Finally, an \emph{oracle} counts a query as correct if SBERT or the Sketch is
correct; it uses the ground truth and is reported only as an upper bound, not as a usable system. Every method returns a single ranking of five candidates, so Top-5 comparisons are fair.

\paragraph{Metrics and statistics.}
We report Top-1, Top-5 and MRR@10. Differences between methods are computed per query (paired), with 95\% confidence intervals obtained by bootstrapping over groups (2{,}000 resamples), since the two queries of a group are correlated.
Across the five seeds, we also report the mean gain, the number of seeds where the interval excludes zero, and the $p$-value of a one-sample $t$-test on the per-seed gains. Because the seeds sample overlapping subsets of the same dataset, they are not fully independent, and the $p$-values should be read as indicative.

\subsubsection{Results}

Table~\ref{tab:main} reports test performance (mean $\pm$ standard deviation over seeds) and
Table~\ref{tab:gains} the paired gains.

\begin{table}[t]
\centering
\caption{Test retrieval performance (\%), mean $\pm$ std over 5 seeds, 7{,}000 test queries per seed.}
\label{tab:main}
\small
\resizebox{\linewidth}{!}{%
\begin{tabular}{lccc}
\toprule
Method & Top-1 & Top-5 & MRR@10 \\
\midrule
SBERT                              & $35.71 \pm 0.59$ & $58.23 \pm 0.96$ & $45.36 \pm 0.59$ \\
Sketch                             & $22.25 \pm 0.82$ & $45.62 \pm 0.82$ & $32.24 \pm 0.78$ \\
Latent (semantic encoder only)     & $23.83 \pm 0.81$ & $46.87 \pm 0.90$ & $33.68 \pm 0.83$ \\
Untrained sketch (control)         & $4.87 \pm 0.23$  & $14.07 \pm 0.16$ & $8.91 \pm 0.18$ \\
\midrule
Threshold cascade SBERT$\to$Sketch & $35.71 \pm 0.59$ & $58.23 \pm 0.96$ & $45.36 \pm 0.59$ \\
Fusion SBERT + Sketch              & $36.30 \pm 0.94$ & $\mathbf{59.59} \pm 0.92$ & $\mathbf{46.23} \pm 0.75$ \\
Fusion SBERT + Latent              & $36.11 \pm 0.70$ & $59.38 \pm 1.06$ & $46.05 \pm 0.66$ \\
Fusion SBERT + untrained sketch    & $35.71 \pm 0.59$ & $58.19 \pm 1.03$ & $45.36 \pm 0.63$ \\
\midrule
Oracle (SBERT or Sketch, upper bound) & $41.37 \pm 0.79$ & $65.55 \pm 1.14$ & $51.69 \pm 0.80$ \\
\bottomrule
\end{tabular}}
\end{table}

\begin{table}[t]
\centering
\caption{Paired gains in points (mean over 5 seeds), with the number of seeds whose 95\% group-bootstrap interval
excludes zero and the $p$-value of the $t$-test on per-seed gains.}
\label{tab:gains}
\footnotesize
\resizebox{\linewidth}{!}{%
\begin{tabular}{lccc}
\toprule
Comparison & Top-1 & Top-5 & MRR@10 \\
\midrule
Fusion (Sketch) $-$ SBERT   & $+0.59$ (2/5, $p{=}0.045$) & $+1.36$ (5/5, $p{<}0.001$) & $+0.87$ (4/5, $p{=}0.007$) \\
Fusion (Latent) $-$ SBERT   & $+0.40$ (1/5, $p{=}0.063$) & $+1.15$ (5/5, $p{<}0.001$) & $+0.69$ (5/5, $p{=}0.006$) \\
Fusion (untrained) $-$ SBERT & $-0.01$ (0/5, $p{=}0.80$) & $-0.04$ (0/5, $p{=}0.54$) & $-0.00$ (0/5, $p{=}0.97$) \\
Cascade $-$ SBERT           & $+0.00$ (0/5) & $+0.00$ (0/5) & $+0.00$ (0/5) \\
Fusion (Sketch) $-$ Fusion (Latent) & $+0.19$ (2/5, $p{=}0.33$) & $+0.21$ (1/5, $p{=}0.12$) & $+0.17$ (2/5, $p{=}0.24$) \\
\bottomrule
\end{tabular}}
\end{table}

\paragraph{SBERT is the strongest single method.}
The Sketch alone loses 13.5 points at Top-1 and 12.6 points at Top-5 with respect to SBERT. The Latent representation, which skips the image bottleneck, is slightly better than the Sketch (+1.6 Top-1, +1.3 Top-5), so the embedding $\to$ sketch $\to$ embedding path loses a small amount of information. At Top-5, the Sketch retains about 97\% of the performance of the Latent.

\paragraph{Score fusion improves over SBERT.}
Fusing SBERT and Sketch scores yields $+1.36$ points at Top-5 (59.59 vs.\ 58.23, i.e.\ $+2.3\%$ relative) and $+0.87$ at MRR@10. The Top-5 gain is positive in all five seeds, with a confidence interval above zero in each of them ($p<0.001$), and the MRR@10 gain is significant in four seeds. At Top-1, the gain ($+0.59$) is only a trend: it is positive in all seeds but significant in two of them, and $p=0.045$ does not survive a Bonferroni correction over the three
metrics (threshold $0.017$). The selected $\alpha$ ranges from 0.30 to 0.65 across seeds, and the test gain as a function of $\alpha$ forms a smooth plateau (about $+1.0$ to $+1.4$ points at Top-5 for $\alpha \in [0.25, 0.60]$, z-score mode), not an isolated peak. The fusion recovers about 19\% of the gap between SBERT and the oracle at Top-5 (1.36 out of 7.32 points) and about 10\% at Top-1.

\paragraph{Controls.}
Fusing SBERT with an untrained sketch model of identical architecture gives no gain ($-0.04$ at Top-5), and the selection procedure correctly falls back to $\alpha \approx 1$ (SBERT) in this case. The improvement is therefore neither an artifact of the hyperparameter selection nor a generic effect of adding an extra projection: it requires a \emph{learned} representation. The threshold cascade does not help either: with the threshold selected on validation, it switches to the Sketch for at most 0.01\% of the queries and is equivalent to SBERT. The oracle shows that the errors of SBERT
and of the Sketch are partly complementary ($+5.7$ Top-1, $+7.3$ Top-5), but this complementarity cannot be exploited by thresholding SBERT's confidence, because low-confidence queries are also hard for the Sketch.

\paragraph{The gain does not come from the image.}
Fusing SBERT with the Latent representation, which never goes through a sketch, yields almost the same gain ($+1.15$ Top-5) as fusing with the Sketch ($+1.36$), and the difference between the two fusions is not significant ($+0.21$ Top-5, $p=0.12$; interval above zero in 1 seed out of 5).

\subsubsection{Conclusions}

\begin{itemize}
  \item A linear fusion of SBERT and sketch-based scores, with $\alpha$ and the normalization chosen on held-out   groups, improves SBERT at Top-5 ($+1.4$ points) and MRR@10 ($+0.9$ points), consistently across seeds and without using   the ground truth. The effect is modest. At Top-1, only a non-robust positive trend is observed ($+0.6$ points).
  \item The gain requires a learned representation (null result for the untrained control) but is
  statistically indistinguishable from the one obtained with the semantic encoder alone. It should therefore be attributed to a model trained on COCO that complements a generic encoder, and \emph{not} to the visual (sketch) representation itself.
  \item A simple similarity-threshold cascade is ineffective, while the oracle upper bound (+5.7 Top-1, +7.3 Top-5) indicates that better combination strategies could, in principle, recover more.
  \item The sketch representation can be defended as an interpretable intermediate that preserves most of the retrieval performance of the latent space (about 97\% at Top-5), but our results provide no evidence that it improves accuracy.
\end{itemize}

\section{Conclusion}

To address the intrinsic opacity of conventional text-embedding models like Sentence-BERT (SBERT), which function as high-dimensional black boxes, this paper has introduced a sketch-based visual grounding framework that transforms caption representations into explicit, visually inspectable topographical sketches. Operating as a structural information bottleneck, the proposed approach prioritizes scene-level composition and spatial layout distributions.

Through comprehensive evaluations on the MS-COCO dataset, several key insights have been established:
\begin{itemize}
    \item \textbf{Retrieval Performance and Trade-offs:} While pure SBERT maintains an advantage in strict top-1 precision and adversarial lexical discrimination, the sketch-based representation closely approaches SBERT performance at top-5 retrieval (retaining approximately 98\%).
    \item \textbf{Score Fusion and Synergy:} Linear score fusion between SBERT and the sketch-based representation yields consistent, modest improvements at Top-5 (+1.4 points) and MRR@10 (+0.9 points) across evaluation seeds. However, control and ablation studies indicate that these performance gains stem from task-specific training rather than the visual nature of the sketch itself, as comparable gains are achieved using the latent representation alone.
    \item \textbf{Visual Transparency:} The sketch representation successfully serves as an interpretable intermediate artifact that provides complete visual transparency into spatial layout encodings while functioning as an effective mechanism for candidate shortlisting.
\end{itemize}

Future work will focus on overcoming the limitations inherited from intermediate SBERT embeddings by developing end-to-end models capable of generating topographical sketches directly from raw text input, thereby investigating their broader utility for advanced reasoning and multi-modal integration.
All the codes used in the experimental section are available in the link given in \cite{Code_Larabi2026} for reproduction.

\end{document}